\documentclass[11pt]{article}
\usepackage[final]{acl}
\usepackage{times}
\usepackage{latexsym}
\usepackage[T1]{fontenc}
\usepackage[utf8]{inputenc}
\usepackage{microtype}
\usepackage{graphicx}
\usepackage{booktabs}
\usepackage{multirow}
\usepackage{amsmath}
\usepackage{amssymb}
\usepackage{xcolor}
\usepackage{array}
\usepackage[table]{xcolor}

\usepackage{pifont}
\newcommand{\cmark}{\ding{51}}
\newcommand{\xmark}{\ding{55}}

\newcommand{\CapLenFull}{20.7}
\newcommand{\CapLenSparseMin}{16.9}
\newcommand{\ChangedMax}{269}
\newcommand{\ChangedMin}{121}
\newcommand{\ChurnMaxPct}{27}

\newcommand{\GenericPctMax}{62}
\newcommand{\GenericPctMin}{48}
\newcommand{\LogfreqShiftMax}{+0.33}
\newcommand{\MaxAbsDeltaPp}{2.5}
\newcommand{\MsrvttN}{998}

\newcommand{\NextqaN}{1000}

\newcommand{\RAtOneFull}{33.0}
\newcommand{\RAtOneWorst}{15.4}
\newcommand{\RetrievalChancePct}{0.10}
\newcommand{\RetrievalPool}{998}
\newcommand{\WorstRatioChanged}{175}
\newcommand{\WorstRatioKeep}{1/4}

\newcommand{\WorstRatioNet}{24}

\newcommand{\camera}[1]{\noindent{#1}}
\title{Generic, Not Wrong: Behavioural Drift under Visual KV-Cache Pruning}

\author{Md Salman Shamil\enspace and\enspace Md Adnan Arefeen \\
  Department of Electrical and Computer Engineering \\
  North South University, Dhaka, Bangladesh \\
  \texttt{\{salman.shamil,adnan.arefeen\}@northsouth.edu}}

\begin{document}
\maketitle

\begin{abstract}
Visual tokens dominate the KV cache of Video-LLMs, making their pruning a common efficiency strategy whose validity is typically established by showing that task accuracy is preserved.
A flat accuracy curve does not imply preserved behaviour: across twelve pruning configurations on NExT-QA, up to 27\% of predictions change while accuracy moves only marginally.
Accuracy also fails to separate uniform-random selection from attention-based criteria, whereas a paired, label-free divergence measure against the model's own full-cache generation separates them with a clear margin.
On MSR-VTT captioning, pruned captions become far less able to identify which video they describe (R@1 \RAtOneFull\% to \RAtOneWorst\%) and more generic in vocabulary, and truncating full-cache captions to the same length does not reproduce this loss.
The reference-based metrics we report show no consistent degradation, tracking caption length instead of content.
We recommend that pruning evaluations report a uniform-random control alongside any attention-based criterion, together with at least one paired, length-robust behavioural measure.
\end{abstract}

\section{Introduction}

\begin{figure*}[t]
\centering
\includegraphics[width=\textwidth]{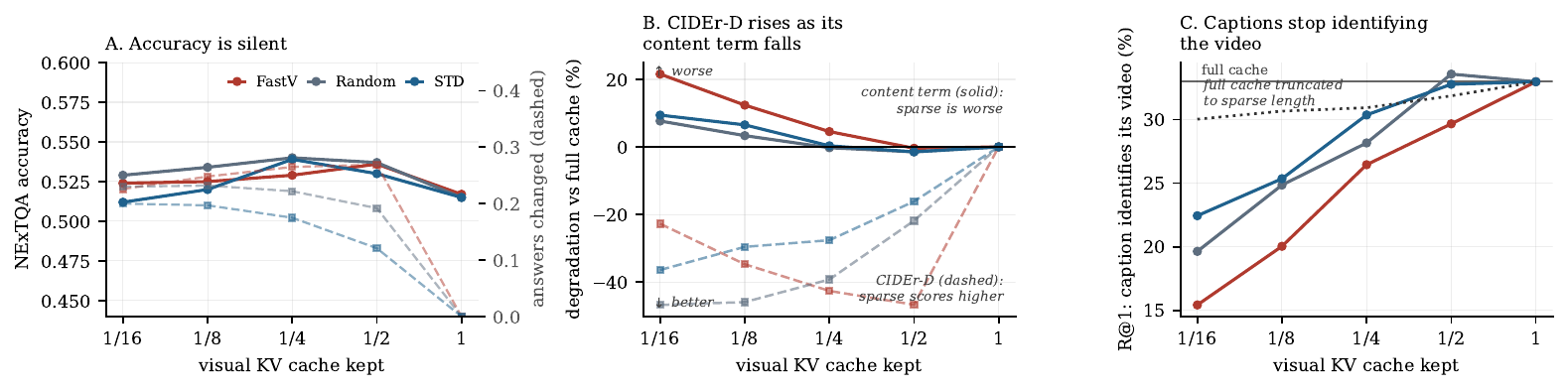}
\caption{\textbf{Two standard instruments miss \camera{behavioural drift}.}
\textbf{(A)} NExT-QA accuracy is flat across a $16\times$ cache reduction (solid, left axis) while as much as \ChurnMaxPct\% of answers differ from the full-cache answer (dashed, right axis).
\textbf{(B)} CIDEr-D scores pruned captions higher (dashed, below zero) while the same score without its length penalty shows them getting worse from $1/4$ onward (solid, above zero); positive is degradation.
\textbf{(C)} Captions identify their own video's reference set less often as the cache shrinks; the dotted line is full-cache captions truncated to the sparse length, so the gap between it and each curve is the loss not explained by length. \camera{FastV and STD denote the FastV-style and STD-style criteria (Section~\ref{sec:setup}).}}
\label{fig:teaser}
\end{figure*}

A Video-LLM that encodes eight frames at 144 tokens each carries over a thousand visual entries in its KV cache before the first answer token is produced, and every decoding step attends to all of them.
Discarding most of that cache is now a standard remedy~\citep{tao2025dycoke}, whether by scoring visual tokens with the attention they receive after an early layer \citep{chen2024image}, restricting that score to attention from the text prompt \citep{zhang2025sparse}, or applying the heavy-hitter and prompt-based selection rules developed for text-only caches \citep{zhang2023h2o,li2024snapkv}.
These criteria differ in what they keep, but they are validated in the same way.
A pruned configuration is accepted when its score on a downstream benchmark stays close to the unpruned model's, and a rule that holds accuracy at a small fraction of the cache is reported as preserving the model's behaviour at that budget.

The inference from a preserved score to preserved behaviour is not merely imprecise but invalid, because accuracy is a sum of per-example indicators and is unchanged by any exchange of errors that leaves their count fixed.
Flat accuracy under aggressive pruning has been reported before \citep{endo2025feather}, and many questions in NExT-QA and other video benchmarks can be answered from a single frame~\citep{buch2022revisiting,lei2023revealing}.
Regardless of the cause, the score cannot certify that the pruned model computes the same function as the dense one.

We therefore measure pruning against the only reference that isolates its effect, the dense model's own output on the same input.
We decode the full-cache generation, then replay it through the pruned cache, advancing both models on the full-cache token at each step so that every comparison is made under an identical prefix, and report the Video Divergence Rate (VDR), the fraction of steps whose top-1 prediction differs, and the First Divergence Time (FDT), the position of the first such step.
Both are label-free and threshold-free, and need only one full-cache generation, shared across all pruned conditions.

We show that
(i) accuracy is preserved by cancellation, moving at most \MaxAbsDeltaPp\ points while up to \ChurnMaxPct\% of NExT-QA predictions change, and random pruning matches attention-based selection;
(ii) reference-based caption metrics on MSR-VTT track generation length and show no consistent degradation, although full-cache captions still score higher at matched length; and
(iii) VDR and FDT register the drift both instruments miss, and length-controlled retrieval and vocabulary measures show captions growing generic, losing the details that distinguish one video from another.

\section{Related Work}
Visual token pruning scores cache entries by early-layer attention or by attention from the text prompt \citep{chen2024image,zhang2025sparse,huang2025prunevid}, and KV-cache pruning for text-only LLMs uses heavy-hitter or observation-window statistics \citep{zhang2023h2o,li2024snapkv}.
Such attention-based scores inherit positional bias from causal masking, and several methods are motivated as corrections to it \citep{endo2025feather,zhang2025beyond}.
\camera{Random selection, which has no such bias, is a strong baseline for visual token pruning in both multimodal LLMs \citep{wen2025token} and video transformers \citep{hao2025principles}.}

Reference-based caption metrics respond to output length as well as content \citep{papineni2002bleu,vedantam2015cider}, which matters because pruning changes generation length.
\camera{For compressed LLMs, \citet{dutta2024accuracy} show that matched accuracy can hide many changed answers, measured as flips and KL divergence.}
\citet{endo2025feather} report that benchmark accuracy is retained under aggressive visual token pruning and turn to localisation tasks to expose the gap, whereas we keep benchmark and selector fixed and compare the pruned model against its own dense output at each decoding step, which needs no labels and no new data.

\section{Measuring Behavioural Drift}\label{sec:setup}
\camera{We take the token-scoring rules from FastV~\citep{chen2024image} and STD~\citep{zhang2025sparse} and compare them, along with random selection, in a single pruning setup.} This section describes the experimental conditions, the selection criteria we compare, and \camera{the measures we use to complement task accuracy.}

\paragraph{Setup.}
We evaluate LLaVA-NeXT-Video-7B \citep{zhang2024llavanextvideo}, sampling 8 frames per video for 1{,}152 visual tokens, and prune the post-prefill visual KV cache to keep ratios of $1$, $1/2$, $1/4$, $1/8$ and $1/16$.
\camera{We use the first question of each of the 1000 videos in the NExT-QA test split \citep{xiao2021next} and 998 MSR-VTT test\_1k clips \citep{xu2016msr}, each with 20 human captions.}\footnote{Two clips were removed by a content filter.}

Decoding is greedy throughout, and every pruned condition is compared against a stored full-cache trace.
The two datasets exercise different generations: on NExT-QA the model produces a multiple-choice answer\camera{\footnote{Options are scored in turn in one context.}} and, separately, a chronology description of up to 128 steps on which divergence is measured, whereas on MSR-VTT divergence is measured on the caption itself, which runs 8--32 steps.

\paragraph{Selection criteria.}
We compare three scoring rules: uniform random selection; a FastV-style score \citep{chen2024image}, the mean attention a visual token receives from all input tokens at layer 2; and an STD-style score \citep{zhang2025sparse}, the same quantity restricted to text tokens.
All three select from the same visual tokens after prefill, prune the visual KV cache to the same size in every layer, and retain every text token, so only the scoring rule varies.
This departs from both systems: FastV prunes tokens during prefill, and STD selects separately for each layer and head and verifies its sparse drafts with the dense model.
We therefore characterise the criteria rather than the systems that introduced them.

\paragraph{Paired divergence.}
We decode the full-cache generation once, then replay it through the pruned cache, forcing the full-cache token at every step so that both models predict from an identical prefix.
This matters because free-running generations separate after their first disagreement, after which any comparison measures divergence of context rather than of computation.
VDR is the mean fraction of steps whose top-1 prediction differs from the full-cache token, and FDT the median position of the first such step as a fraction of the generation, both label-free, threshold-free, and obtained from one forward pass over the pruned model.
\camera{VDR measures how often pruning changes the model's next-token decision, and FDT how soon the first change occurs. Unlike a count of changed answers, VDR also applies to open-ended output such as captions, and unlike the KL divergence between the two output distributions, it counts decisions, which are easy to interpret. We report VDR alongside accuracy, with FDT as a secondary diagnostic. Because both models see the same prefix, VDR is a local measure, so on MSR-VTT we also compare the free-running captions themselves by the Jaccard overlap of their token types.}

\paragraph{Length-robust caption measures.}
Like \citet{wang2020compare}, we measure how distinctive a caption is by comparing it with the reference captions of other inputs.
On MSR-VTT we build one document per video from its pooled references and score each caption against that fixed corpus by TF--IDF cosine over unigrams and bigrams, reporting R@1 among \RetrievalPool\ documents against \RetrievalChancePct\% chance.
As a length control we also rank full-cache captions truncated to the sparse caption's length.
The same length concern applies to the reference-based metrics: CIDEr-D combines an n-gram content score with a Gaussian penalty on the difference between candidate and reference length \citep{vedantam2015cider}, and we report the content score with that penalty disabled.
We separately track the mean log document frequency of the words a caption uses, measured over the full-cache captions, and the rate of generic scene, colour and clothing vocabulary.

\begin{table}[!htbp]\centering\small
\setlength{\tabcolsep}{3pt}
\begin{tabular}{lcccccc}
\toprule
\multirow{2}{*}{Method} & Keep & \multirow{2}{*}{Acc.} & Answer & C$\to$W & \multirow{2}{*}{VDR} & \multirow{2}{*}{FDT} \\
 & Ratio & & Changes & /W$\to$C & & \\
\midrule
Full cache & $1$ & 0.515 & 0.0\% & 0/0 & 0.000 & -- \\
\midrule
\multirow{4}{*}{FastV} & $1/2$ & 0.536 & 26.9\% & 79/98 & 0.058 & 11\% \\
 & $1/4$ & 0.529 & 26.5\% & 80/92 & 0.082 & 9\% \\
 & $1/8$ & 0.525 & 24.8\% & 74/82 & 0.101 & 8\% \\
 & $1/16$ & 0.524 & 22.6\% & 67/74 & 0.117 & 7\% \\
\midrule
\multirow{4}{*}{Random} & $1/2$ & 0.537 & 19.2\% & 52/74 & 0.050 & 13\% \\
 & $1/4$ & 0.540 & 22.2\% & 61/86 & 0.081 & 9\% \\
 & $1/8$ & 0.534 & 23.2\% & 65/84 & 0.106 & 8\% \\
 & $1/16$ & 0.529 & 23.0\% & 66/80 & 0.124 & 7\% \\
\midrule
\multirow{4}{*}{STD} & $1/2$ & 0.530 & 12.1\% & 31/46 & 0.036 & 16\% \\
 & $1/4$ & 0.539 & 17.5\% & 46/70 & 0.068 & 10\% \\
 & $1/8$ & 0.520 & 19.7\% & 62/67 & 0.097 & 8\% \\
 & $1/16$ & 0.512 & 20.0\% & 66/63 & 0.118 & 7\% \\
\bottomrule
\end{tabular}
\caption{NExT-QA under visual KV-cache pruning ($N{=}\NextqaN$). Accuracy stays within \MaxAbsDeltaPp\ percentage points of full cache in every condition while \ChangedMin--\ChangedMax\ predictions change, because the correct-to-wrong (C$\to$W) and wrong-to-correct (W$\to$C) flips nearly cancel. Decoding is greedy and every condition is compared against a stored full-cache trace, so the full-cache row differs from itself nowhere by construction. VDR and FDT are measured on a free-form description of the video, generated separately from the answer. VDR is the mean fraction of steps whose top-1 token differs from full cache, and FDT the median position of the first such step as a fraction of the generation.}
\label{tab:churn}
\end{table}

\section{Results}
Table~\ref{tab:churn} reports NExT-QA and Table~\ref{tab:caption} MSR-VTT, in both cases for the three selection criteria at four pruned budgets against the full-cache model, and Figure~\ref{fig:teaser} summarises the key results.

\begin{table*}[t]
\centering\small
\setlength{\tabcolsep}{5pt}
\begin{tabular}{llrcccccc}
\toprule
Model & Dataset & $N$ & \shortstack{Full-cache\\Accuracy} & \shortstack{$1/8$\\Accuracy} & \shortstack{Answer\\Changes} & C$\to$W / W$\to$C & VDR & FDT \\
\midrule
LLaVA-NeXT-Video-7B & NExT-QA     & 1000 & 0.515 & 0.520 & 19.7\% & 62 / 67 & 0.097 & 8\% \\
\midrule
LLaVA-NeXT-Video-7B & NExT-QA     & 100  & 0.450 & 0.420 & 24.0\% & 9 / 6   & 0.096 & 7\% \\
SmolVLM2-2.2B       & NExT-QA     & 100  & 0.350 & 0.370 &  9.0\% & 2 / 4   & 0.107 & 5\% \\
LLaVA-NeXT-Video-7B & TempCompass & 100  & 0.440 & 0.480 & 24.0\% & 6 / 10  & 0.103 & 7\% \\
\bottomrule
\end{tabular}
\caption{\camera{Results at $1/8$ cache with the STD-style criterion. The first row repeats the full NExT-QA result from Table~\ref{tab:churn}; the second runs the same setup on 100 questions and gives nearly the same VDR and FDT. The last two rows change the model and the benchmark.}}
\label{tab:checks}
\end{table*}

\begin{figure}[!tbp]
\centering
\includegraphics[width=\columnwidth]{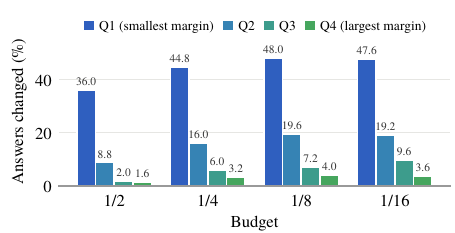}
\caption{\camera{Share of NExT-QA answers that change under STD-style pruning, within each of four equal groups of 250 questions. The groups are sorted by the gap between the full-cache model's two highest option scores, so Q1 holds the questions where it was closest to picking another option. The groups stay the same at every budget.}}
\label{fig:margin}
\end{figure}

\paragraph{Accuracy is preserved by cancellation.}
Accuracy is retained in every pruned configuration, staying within \MaxAbsDeltaPp\ percentage points of the full-cache model.
The predictions underneath do not hold still: between \ChangedMin\ and \ChangedMax\ of \NextqaN\ answers change, and the two directions almost annihilate, with \mbox{\camera{STD-style}} at $\WorstRatioKeep$ changing \WorstRatioChanged\ predictions for a net of \WorstRatioNet.
\camera{
Answer changes are most frequent when the full-cache model's top two option scores are close, but they also occur when the gap is large (Figure~\ref{fig:margin}).
The same cancellation appears with SmolVLM2-2.2B~\citep{marafioti2025smolvlm} on NExT-QA and with LLaVA-NeXT-Video on TempCompass~\citep{liu2024tempcompass}, where 9--24\% of answers change at 1/8 cache while accuracy moves by at most 4 points (Table~\ref{tab:checks}).
}
In Table~\ref{tab:churn}, accuracy does not fall; it drifts upward, which we read as insensitivity to pruning.
It also fails to separate the criteria, with seeded random selection never the worst of the three at any budget.
\camera{Earlier work finds random selection competitive with existing pruning methods in other settings \citep{wen2025token,hao2025principles}.}
Unlike accuracy, VDR separates the criteria, reading 0.036 for STD-style, 0.050 for Random and 0.058 for FastV-style at $1/2$ on the model's video description, though the three converge near 0.12 at $1/16$.\footnote{All pairwise differences at $1/2$ are significant (Wilcoxon, $p<0.001$).}
\camera{Mean $\mathrm{KL}(p_{\text{full}} \,\|\, p_{\text{pruned}})$ over next-token distributions gives the same ordering at $1/2$, with 0.009 for STD-style, 0.012 for Random and 0.019 for FastV-style.\footnote{At each step, KL is computed over the union of the 256 most probable tokens under each model, with the rest of each distribution merged into one extra outcome.}}

\begin{table}[t]\centering\small
\setlength{\tabcolsep}{2.5pt}
\begin{tabular}{lcccccc}
\toprule
\multirow{2}{*}{Method} & Keep & CIDEr & \multirow{2}{*}{cont.} & \multirow{2}{*}{Agr.} & \multirow{2}{*}{Gen.} & \multirow{2}{*}{R@1} \\
 & Ratio & -D & & & & \\
\midrule
Full cache & 1 & 0.142 & 0.402 & 1.000 & 0.0\% & 33.0\% \\
\midrule
\multirow{4}{*}{FastV} & $1/2$  & 0.208 & 0.403 & 0.519 & +14\% & 29.7\% \\
 & $1/4$  & 0.202 & 0.383 & 0.457 & +25\% & 26.5\% \\
 & $1/8$  & 0.191 & 0.352 & 0.413 & +44\% & 20.0\% \\
 & $1/16$ & 0.174 & 0.315 & 0.381 & +62\% & 15.4\% \\
\midrule
\multirow{4}{*}{Random} & $1/2$  & 0.173 & 0.407 & 0.656 & +12\% & \textbf{33.6\%} \\
 & $1/4$  & 0.198 & 0.403 & 0.546 & +22\% & 28.2\% \\
 & $1/8$  & 0.207 & 0.388 & 0.473 & +32\% & 24.9\% \\
 & $1/16$ & 0.208 & 0.371 & 0.423 & +49\% & 19.6\% \\
\midrule
\multirow{4}{*}{STD} & $1/2$  & 0.165 & 0.407 & 0.732 & +4\%  & 32.8\% \\
 & $1/4$  & 0.181 & 0.400 & 0.579 & +14\% & \textbf{30.4\%} \\
 & $1/8$  & 0.184 & 0.375 & 0.482 & +28\% & \textbf{25.4\%} \\
 & $1/16$ & 0.194 & 0.364 & 0.425 & +48\% & \textbf{22.4\%} \\
\bottomrule
\end{tabular}
\caption{MSR-VTT captioning under visual KV-cache pruning ($N{=}\MsrvttN$). CIDEr-D rises under pruning while the same score computed without its Gaussian length penalty (cont.)\ is flat at $1/2$ and falls thereafter, as pruning shortens captions toward the reference length. Agr.\ is the Jaccard overlap between the token types of the pruned and full-cache captions, so the full-cache row is 1 by definition. Gen.\ is the relative change in the rate of generic scene vocabulary. R@1 is how often a caption ranks its own video's pooled reference set first among \RetrievalPool\ such sets by TF--IDF cosine (chance \RetrievalChancePct\%). Bold marks the highest R@1 at each budget.}
\label{tab:caption}
\end{table}

\paragraph{Standard caption metrics do not register the change.}
On MSR-VTT the pruned model writes shorter captions, \CapLenFull\ tokens falling to \CapLenSparseMin\ against references averaging 9.3, and the standard metrics show no consistent loss due to pruning: CIDEr-D moves from 0.142 to 0.208 (Table~\ref{tab:caption}), while ROUGE-L~\citep{lin2004rouge}, BLEU-4~\citep{papineni2002bleu} and METEOR~\citep{banerjee2005meteor} show no consistent degradation.\footnote{ROUGE-L rises in all twelve conditions and BLEU-4 in eleven; METEOR rises at the two mildest budgets and falls at the two most aggressive.}
Truncated to the same length, full-cache captions score higher on BLEU-4 and ROUGE-L in every condition, by 15\% and 4\% for STD-style at $1/16$.
CIDEr-D's content score agrees, flat at $1/2$ and falling to 0.315 for FastV-style at $1/16$, as does token overlap with the model's full-cache caption, which drops to 0.381.

\begin{figure}[!bp]
\centering
\includegraphics[width=\columnwidth]{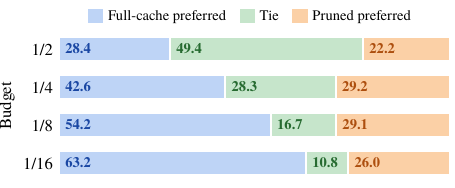}
\caption{\camera{Blinded pairwise LLM judge on MSR-VTT ($N{=}998$, STD-style): percentage of clips where the judge prefers each caption.}}
\label{fig:judge}
\end{figure}

\definecolor{qred}{rgb}{0.72,0.11,0.11}
\definecolor{qgreen}{rgb}{0.00,0.45,0.20}
\newcommand{\qr}[1]{\textcolor{qred}{#1}}
\newcommand{\qg}[1]{\textcolor{qgreen}{#1}}
\begin{table*}[t]\centering
\begin{tabular}{@{}ll@{}}
\begin{minipage}[t]{0.47\textwidth}
\includegraphics[width=\linewidth]{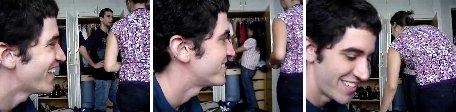}\\[1pt]
\footnotesize\baselineskip=8pt
\textit{Q:} where are the people hanging out\\
\textit{Full:} \qg{bedroom~\cmark}\quad\textit{STD $1/8$:} \qr{water fountain~\xmark}\\
\end{minipage} &
\begin{minipage}[t]{0.47\textwidth}
\includegraphics[height=1.82cm,width=\linewidth]{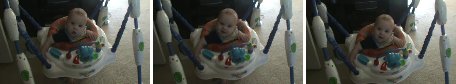}\\[1pt]
\footnotesize\baselineskip=8pt
\textit{Q:} how did the baby rest his head a few times in the video\\
\textit{Full:} \qr{with his hands~\xmark}\quad\textit{STD $1/8$:} \qg{on the cushion~\cmark}\\
\end{minipage} \\[4pt]
\begin{minipage}[t]{0.47\textwidth}
\includegraphics[width=\linewidth]{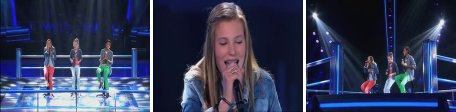}\\[1pt]
\footnotesize\baselineskip=8pt
\textit{Ref:} some teenagers are singing on stage and sitting on stools\\
\textit{Full:} A young woman singing into a microphone on \qr{stage,} with a man and a \qr{girl} singing \qr{behind} her.\\
\textit{STD $1/8$:} A young woman singing into a microphone \qr{while} a man \qr{watches.}\\
\scriptsize CIDEr-D: 0.127$\to$0.163\quad Rank: 19$\to$32
\end{minipage} &
\begin{minipage}[t]{0.47\textwidth}
\includegraphics[width=\linewidth]{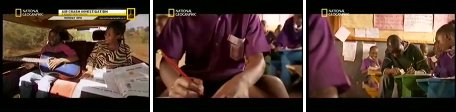}\\[1pt]
\footnotesize\baselineskip=8pt
\textit{Ref:} guy in purple tshirt playing guitar as they drive through\\
\textit{Full:} A man \qr{plays} guitar in a \qr{car} while a woman \qr{listens,} and a \qr{group} of \qr{children} are \qr{seen} in a \qr{classroom.}\\
\textit{STD $1/8$:} A man is \qr{playing} guitar in a \qr{room} while a woman is \qr{singing.}\\
\scriptsize CIDEr-D: 0.213$\to$0.440\quad Rank: 4$\to$25
\end{minipage} \\[4pt]
\end{tabular}
\caption{Examples at $1/8$ cache with the \camera{STD-style criterion}. \textbf{Top (NExT-QA):} one answer that became wrong and one that became correct under the same pruning, the cancellation that keeps accuracy flat. \textbf{Bottom (MSR-VTT):} the pruned caption drops concrete detail while CIDEr-D \emph{rises}, as the shorter caption sits closer to the reference length, and the reference set's rank falls at the same time. Red marks content words present in one caption but not the other. Each strip shows three downscaled frames from the clip.}
\label{tab:qualitative}
\end{table*}

\paragraph{The captions lose what identifies the video.}
Retrieval falls with the budget, from \RAtOneFull\% R@1 at full cache to \RAtOneWorst\% at $1/16$, and the median rank of the correct reference set rises from 3 to 17.
Length does not account for this: full-cache captions truncated to the sparse length still retrieve at roughly 30\% (Figure~\ref{fig:teaser}C), so truncation costs about 3 points against pruning's 17.6.
The words themselves get more common (mean log frequency $\LogfreqShiftMax$) and more generic (up \GenericPctMin--\GenericPctMax\% at $1/16$), while colour and clothing terms fall by 36--39\% and 39--42\%.
A caption can only lose rank by dropping words its references contain, and the share of caption content matching no reference stays at 47.7\%, so the loss is missing detail and not new error; Table~\ref{tab:qualitative} shows two captions and two answers at $1/8$.

\paragraph{LLM judge.}
\camera{We also compared full-cache and pruned captions for all 998 MSR-VTT clips with GPT-5.6 Luna under STD-style pruning, using a blinded pairwise judge~\citep{zheng2023judging}. The judge saw the reference captions and the two candidates in random order~\citep{wang2024large}, without knowing which was pruned, and chose the one that better preserved the content and detail of the video, or a tie. Its preference for the full-cache caption grows as the budget shrinks (Figure~\ref{fig:judge}), following the fall in R@1 that the standard reference-based metrics do not show.}

\section{Conclusion}
Visual KV-cache pruning changes what a Video-LLM says while leaving the scores used to validate it largely unmoved.
On NExT-QA, accuracy holds because errors cancel, and it does not separate attention-based selection from a random control.
On MSR-VTT the reference-based caption metrics move with generation length, and at matched length the full-cache captions still score higher.
Comparing the pruned model against its own dense output does show the loss, in captions that get more generic and identify their video less often.
None of this shows that pruning is harmful, only that the standard instruments cannot see what it does.
If the goal of pruning is a model that behaves like the one it replaces, then behaviour is what has to be measured, because a benchmark score can hold even when behaviour changes.
We therefore recommend that pruning evaluations report a uniform-random control alongside any attention-based criterion, VDR alongside accuracy, and, for captions, at least one length-robust measure such as R@1.

\section*{Use of Generative AI}
The authors used generative AI tools to improve fluency and clarity of the writing and for limited coding support.
All scientific content is the work of the human authors, who take full responsibility for the paper.

\section*{Limitations}
Our results characterise the selection criteria themselves, not the systems that introduced them, and do not transfer to FastV or STD in their native deployments.
In particular, STD is lossless in its original setting, where a dense model verifies each draft.
Our main experiments use one model at one scale, frame count and precision under greedy decoding, and we do not claim that the magnitudes reported here transfer to all settings.
\camera{VDR isolates each prediction by using the full-cache prefix, so it does not follow free-running drift, and it counts minor rewordings as changes. We report the free-running MSR-VTT measures and the LLM judge alongside it to address both.}
The genericity measures rely in part on fixed word lists, which is why we report the corpus log-frequency shift alongside them as a list-free measure.

\bibliography{custom}
\end{document}